# Predicting the Type and Target of Offensive Social Media Posts in Marathi

Marcos Zampieri[1*], Tharindu Ranasinghe[2], Mrinal Chaudhari[1], Saurabh Gaikwad[1], Prajwal Krishna[1], Mayuresh Nene[1] and Shrunali Paygude[1]

[1]Rochester Institute of Technology, Rochester, NY, USA.
[2]University of Wolverhampton, Wolverhampton, UK.

*Corresponding author(s). E-mail(s): marcos.zampieri@rit.edu;

**Abstract**

The presence of offensive language on social media is very common motivating platforms to invest in strategies to make communities safer. This includes developing robust machine learning systems capable of recognizing offensive content online. Apart from a few notable exceptions, most research on automatic offensive language identification has dealt with English and a few other high resource languages such as French, German, and Spanish. In this paper we address this gap by tackling offensive language identification in Marathi, a low-resource Indo-Aryan language spoken in India. We introduce the Marathi Offensive Language Dataset v.2.0 or *MOLD 2.0* and present multiple experiments on this dataset. *MOLD 2.0* is a much larger version of *MOLD* with expanded annotation to the levels B (type) and C (target) of the popular OLID taxonomy. *MOLD 2.0* is the first hierarchical offensive language dataset compiled for Marathi, thus opening new avenues for research in low-resource Indo-Aryan languages. Finally, we also introduce *SeMOLD*, a larger dataset annotated following the semi-supervised methods presented in SOLID [1].

# 1 Introduction

The widespread of offensive content online such as hate speech and cyberbullying is a global phenomenon. This has sparked interest in the AI and NLP communities motivating the development of various systems trained to automatically detect potentially harmful content [2]. Even though thousands of languages and dialects are widely used in social media, the clear majority of these studies consider English only. This is evidenced by the creation of many offensive language resources for English such as annotated datasets [1], lexicons [3], and pre-trained models [4].

More recently researchers have turned their attention to the problem of offensive content in other languages such as Arabic [5], French [6], Greek [7], and Portuguese [8], to name a few. In doing so, they have created new datasets and resources for each of these languages. Competitions such as OffensEval [9] and TRAC [10] provided multilingual datasets compiled and annotated using the same methodology. The availability of multilingual has made it possible to explore data augmentation methods [11], multilingual word embeddings [12], and cross-lingual contextual word embeddings [13].

In this paper, we revisit the task of offensive language identification for low resource languages, that is, languages for which few or no corpora, datasets, and language processing tools are available. Our work focus on Marathi, an Indo-Aryan language spoken by over 80 million people, most of whom live in the Indian state of Maharashtra. Even though Marathi is spoken by a large population, it is relatively low-resourced compared to other languages spoken in the region, most notably Hindi, the most similar language to Marathi. We collect and annotate data from Twitter to create the largest Marathi offensive language identification dataset to date. Furthermore, we train a number of state-of-the-art computational models on this dataset and evaluate the results in detail which makes this paper the first comprehensive evaluation on Marathi offensive language online.

This paper presents the following contributions:

1. We release MOLD 2.0[1], the largest annotated Marathi Offensive Language Dataset to date. MOLD 2.0 contains more than 3,600 annotated tweets annotated using the popular OLID [14] three-level hierarchical annotation schema; (A) Offensive Language Detection (B) Categorization of Offensive Language (C) Offensive Language Target Identification.
2. We experiment with several machine learning models including state-of-the-art transformer models to predict the type and target of offensive tweets in Marathi. To the best of our knowledge, the identification of types and targets of offensive posts have not been attempted on Marathi.
3. We explore offensive language identification with cross-lingual embeddings and transfer learning. We take advantage of existing data in high-resource languages such as English and Hindi, to project predictions to Marathi.

---

[1]Dataset available at: https://github.com/tharindudr/MOLD

   We show that transfer learning can improve the results on Marathi which could benefit a multitude of low-resource languages.
4. Finally, we investigate semi-supervised data augmentation. We create *SeMOLD*, a larger semi-supervised dataset with more than 8,000 instances for Marathi. We use multiple machine learning models trained on the annotated training set and combine the scores following a similar methodology described in [1]. We show that this semi-supervised dataset can be used to augment the training set which leads to improves results of machine learning models.

The development MOLD 2.0 and SeMOLD open exciting new avenues for research in Marathi offensive language identification. With these two resources, we aim to answer the following research questions:

- **RQ1**: To which extent is it possible to identify types and targets of offensive posts in Marathi?
- **RQ2**: Our second research question addresses data scarcity, a known challenge for low-resource NLP. We divide it in two parts as follows:
  - **RQ2.1**: How does data size influences performance in Marathi offensive language identification?
  - **RQ2.2**: Do available resources from resource-rich languages combine with transfer-learning techniques aid the identification of types and targets in Marathi offensive language language identification?

Previous work [15] has addressed the identification of offensive posts in Marathi but the types and targets included in offensive posts, the core part of the popular OLID taxonomy [14], have not been addressed for Marathi. Finally, with respect to data size and transfer learning, we draw inspiration on recent work that applied cross-lingual models for low-resource offensive language identification [13, 16] applying it to Marathi.

## 2 Related Work

The problem of offensive content online continues to attract attention within the AI and NLP communities. In recent studies, researchers have developed systems to identify whether a post or part thereof is considered offensive [17] or to predict whether conversations will go awry [18]. Popular international competitions on the topic have been organized at conferences such as HASOC [19, 20], HatEval [21], OffensEval [22], and TRAC [23, 24]. These competitions attracted a large number of participants and they provided participants with various of important benchmark datasets.

A variety of computing models have been proposed to tackle offensive content online ranging from classical machine learning classifiers such as SVMs with feature engineering [25, 26] to deep neural networks combined with word embeddings [27, 28]. With the recent development of large pre-trained transformer models such as BERT and XLNET [29, 30], several studies have explored the use of general pre-trained transformers [31, 32] while others have worked on fine-tuning models on offensive language corpora such as fBERT [4].

In terms of languages, due to the availability of suitable datasets, the vast majority of studies in offensive language identification use English data [2, 33]. In the past few years, however, more offensive language dataset have been for languages other than English such as Arabic [5], Dutch [34], French [6], German [35], Greek [7], Italian [36], Portuguese [8], Slovene [37], Turkish [38], and many others. To the best of our knowledge, the only Marathi dataset available to date is the aforementioned Marathi Offensive Language Dataset (MOLD) [15], a manually annotated dataset containing nearly 2,500 tweets. Our work builds on MOLD by applying the same data collection methods to expand it in terms of both size and annotation.

Finally, multilingual offensive language identification is a recent trend that takes advantage of large pre-trained cross-lingual and multilingual models such as XLM-R [39]. Using this architecture, it is possible to leverage available English resources to make predictions in languages with less resources helping to cope with data scarcity in low-resource languages [13, 17].

# 3 Data Collection

MOLD 2.0 builds on the research presented in [15] which introduced MOLD 1.0. The annotation of both MOLD 1.0 and MOLD 2.0 follows the OLID annotation taxonomy which includes three levels (labels in brackets):

- **Level A:** Offensive (OFF)/Non-offensive (NOT).
- **Level B:** Classification of the type of offensive (OFF) tweet - Targeted (TIN)/Untargeted (UNT).
- **Level C:** Classification of the target of a targeted (TIN) tweet - Individual(IND)/Group(GRP) or Other(OTH).

Our initial dataset (MOLD 1.0) consisted of nearly 2,500 tweets. As shown in Table 1, we collected 1,100 additional instances for MOLD 2.0 resulting in a dataset of 3,611 tweets according to the same methodology described in [15]. Data collection was carried out with a data extraction script which utilized the Tweepy[2] library along with the API provided by Twitter.

As MOLD 1.0 was only annotated on OLID Level A, in MOLD 2.0 we expand the annotation to the full three-level OLID taxonomy annotating Level B and Level C. Examples from the dataset along with English translation are presented in Table 2. The annotation was carried out by the 3 native speakers of Marathi. The annotators were a mix of male (1) and female (2) Master's students working in the project. We provided the annotators with guidelines on how to annotate the data and supervised the process with periodic meetings to make sure they were correctly following the guidelines.

---
[2]Tweeypy Python library documentation is available on https://www.tweepy.org/

| A | B | C | Training | Test | Total |
|---|---|---|---|---|---|
| OFF | TIN | IND | 503 | 51 | 554 |
| OFF | TIN | OTH | 80 | 56 | 136 |
| OFF | TIN | GRP | 157 | 51 | 208 |
| OFF | UNT | — | 327 | 102 | 429 |
| NOT | — | — | 2,034 | 250 | 2,284 |
| **All** | | | 3,101 | 510 | 3,611 |

**Table 1** MOLD v2.0 - Distribution of label combinations.

We report an inter-annotator agreement of 0.79 Cohen's kappa [40] on the three levels.

| Tweet | A | B | C |
|---|---|---|---|
| कोण आहे तुमचा आवडता? (Who is your favorite?) | NOT | — | — |
| बावळट ना मग काय अजूनं (Stupid, what else?) | OFF | UNT | — |
| रंडी साली. (Damn slut) | OFF | TIN | IND |
| दळभद्री हरामखोरांचे सरकार आहे हे. (This is a government of thankless sick-heads.) | OFF | TIN | GRP |
| भिकारचोट लेकाचे मिडिया कलाकार सगळे विकले गेले आहे राज्य सरकारच्या कामासाठी. (These artists and media without standards have sold themselves to work for the state government.) | OFF | TIN | OTH |

**Table 2** Four tweets from the dataset, with their labels for each level of the annotation schema. English translations are inside brackets

Finally, following the same methodology described for MOLD 2.0, we collected an additional 8,000 instances from Twitter to create SeMOLD, a larger dataset with semi-supervised annotation presented in Section 6.

## 4 Experiments and Evaluation

We experimented with several machine learning models trained on the training set, and evaluated by predicting the labels for the held-out test set. As the label distribution is highly imbalanced, we evaluate and compare the performance of the different models using macro-averaged F1-score. We further report per-class Precision (P), Recall (R), and F1-score (F1), and weighted average. Finally, we compare the performance of the models against simple majority and minority class baselines.

### SVC

Our simplest machine learning model is a linear Support Vector Classifier (SVC) trained on word unigrams. Before the emergence of neural networks, SVCs have achieved state-of-the-art results for many text classification tasks [41, 42] including offensive language identification [14, 43]. Even in the neural network era SVCs produce an efficient and effective baseline.

### BiLSTM

As the first embedding-based neural model we experimented with a bidirectional Long Short-Term-Memory (BiLSTM) model, which we adopted from a pre-existing model for Greek offensive language identification [7]. The model consists of (i) an input embedding layer, (ii) two bidirectional LSTM layers, and (iii) two dense layers. The output of the final dense layer is ultimately passed through a softmax layer to produce the final prediction. The architecture diagram of the BiLSTM model is shown in Figure 1. Our BiLSTM layer has 64 units while the first dense layer had 256 units.

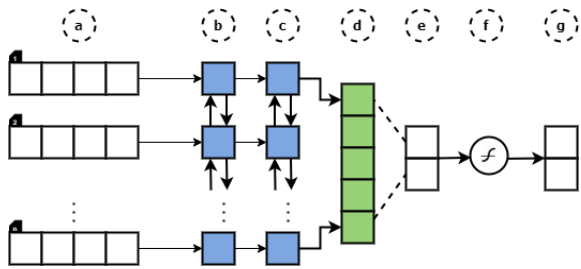

**Figure 1** The BiLSTM model for Marathi offensive language identification. The labels are **(a)** input embeddings, **(b,c)** two BiLSTM layers, **(d, e)** fully-connected layers; **(f)** softmax activation, and **(g)** final probabilities.

### CNN

We also experimented with a convolutional neural network (CNN), which we adopted from a pre-existing model for English sentiment classification [44]. The model consists of (i) an input embedding layer, (ii) 1 dimensional CNN layer (1DCNN), (iii) max pooling layer and (iv) two dense layers. The output of the final dense layer is ultimately passed through a softmax layer to produce the final prediction.

For the BiLSTM and CNN models presented above, we set three input channels for the input embedding layers: pre-trained Marathi FastText embeddings[3] [45], Continuous Bag of Words Model for Marathi[4] [46] as well as updatable embeddings learned by the model during training.

---

[3] Marathi FastText embeddings are available on https://fasttext.cc/docs/en/crawl-vectors.html
[4] Marathi word embeddings are available on https://www.cfilt.iitb.ac.in/~diptesh/embeddings/

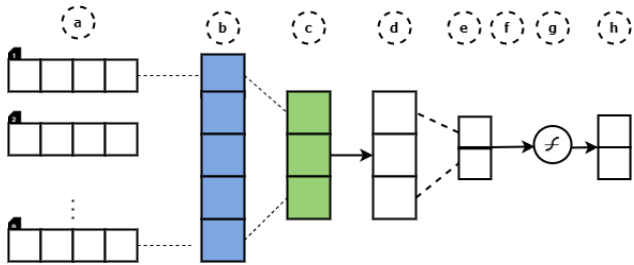

**Figure 2** CNN model for Marathi offensive language identification. The labels are **(a)** input embeddings, **(b)** 1DCNN, **(c)** max pooling, **(d, e)** fully-connected layer; **(f)** with dropout, **(g)** softmax activation, and **(h)** final probabilities.

### Transformers

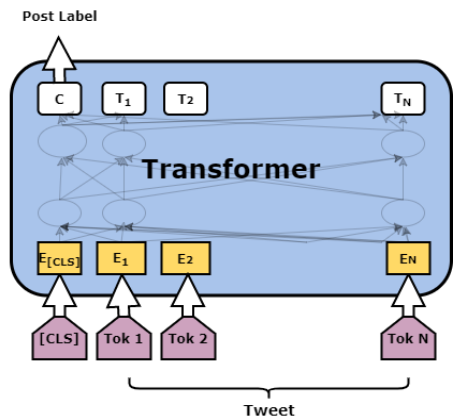

**Figure 3** Transformer model for Marathi offensive language identification [13]

Finally we experimented with several pre-trained transformer models. With the introduction of BERT [29], transformer models have achieved state-of-the-art performance in many natural language processing tasks [29] including offensive language identification [4, 13, 17, 47]. From an input sentence, transformers compute a feature vector $\boldsymbol{h} \in \mathbb{R}^d$, upon which we build a classifier for the task. For this task, we implemented a softmax layer, i.e., the predicted probabilities are $\boldsymbol{y}^{(B)} = \text{softmax}(W\boldsymbol{h})$, where $W \in \mathbb{R}^{k \times d}$ is the softmax weight matrix and $k$ is the number of labels. In our experiments, we used three pre-trained transformer models available in HuggingFace model hub [48] that supports Marathi; mBERT [29], xlm-roberta-large [39] and IndicBERT [49]. The implementation was adopted from the *DeepOffense* Python library[5]. The overall transformer architecture is available in Figure 3.

---

[5]DeepOffense is available as a pip package in https://pypi.org/project/deepoffense/

| Type | Model | OFF | | | NOT | | | Weighted | | | Macro F1 |
|---|---|---|---|---|---|---|---|---|---|---|---|
| | | P | R | F1 | P | R | F1 | P | R | F1 | |
| Traditional | SVM | 0.72 | 0.68 | 0.70 | 0.84 | 0.80 | 0.82 | 0.80 | 0.76 | 0.78 | 0.74 |
| BiLSTM | CBOW | 0.75 | 0.71 | 0.73 | 0.87 | 0.83 | 0.85 | 0.83 | 0.81 | 0.80 | 0.77 |
| | fastText | 0.75 | 0.72 | 0.74 | 0.88 | 0.83 | 0.86 | 0.84 | 0.82 | 0.81 | 0.78 |
| | Self-learned | 0.73 | 0.69 | 0.71 | 0.85 | 0.81 | 0.83 | 0.79 | 0.76 | 0.77 | 0.76 |
| CNN | CBOW | 0.77 | 0.73 | 0.75 | 0.89 | 0.85 | 0.86 | 0.85 | 0.83 | 0.82 | 0.80 |
| | fastText | 0.78 | 0.74 | 0.76 | 0.90 | 0.86 | 0.87 | 0.87 | 0.86 | 0.83 | 0.81 |
| | Self-learned | 0.76 | 0.72 | 0.74 | 0.88 | 0.83 | 0.84 | 0.83 | 0.81 | 0.82 | 0.80 |
| Transformers | mBERT | 0.79 | 0.75 | 0.77 | 0.91 | 0.87 | 0.88 | 0.88 | 0.86 | 0.84 | 0.82 |
| | XLM-R | 0.81 | 0.77 | 0.79 | 0.93 | 0.89 | 0.90 | 0.90 | 0.88 | 0.86 | 0.84 |
| | IndicBERT | 0.83 | 0.79 | 0.81 | 0.95 | 0.91 | 0.91 | 0.91 | 0.89 | 0.88 | **0.85** |
| Baseline | All OFF | 1.00 | 0.51 | 0.67 | 0.00 | 0.00 | 0.00 | 1.00 | 0.51 | 0.67 | 0.33 |
| | All NOT | 0.00 | 0.00 | 0.00 | 1.00 | 0.49 | 0.65 | 1.00 | 0.49 | 0.65 | 0.33 |

**Table 3** Results for offensive language detection (level A). We report Precision (P), Recall (R), and F1 for each model/baseline on all classes (OFF, NOT), and weighted averages. Macro-F1 is also listed (best in bold).

For the transformer-based models, we employed a batch-size of 16, Adam optimiser with learning rate 2e−5, and a linear learning rate warm-up over 10% of the training data. During the training process, the parameters of the transformer model, as well as the parameters of the subsequent layers, were updated. The models were evaluated while training using an evaluation set that had one fifth of the rows in training data. We performed early stopping if the evaluation loss did not improve over three evaluation steps. All the models were trained for three epochs.

### *Offensive Language Detection*

The performance on discriminating between offensive (OFF) and non-offensive (NOT) posts is reported in Table 3. We can see that all models perform better than the majority baseline. As expected, transformer-based models outperform other machine learning models. From the transformer models, IndicBERT model [49] outperforms general multilingual transformer models such as mBERT [29] and xlm-roberta-large [39] providing 0.85 Macro F1 score on the test set.

### *Categorization of Offensive Language*

In this set of experiments, the models were trained to discriminate between targeted insults and threats (TIN) and untargeted (UNT) offenses. The performance of various machine learning models on this task is shown in Table 4. Similar to level A, transformer models outperformed other machine learning models in this set of experiments too. Furthermore, IndicBERT model [49] performs best from the transformer model with providing 0.74 Macro F1 score.

| Type | Model | TIN | | | UNT | | | Weighted | | | Macro F1 |
|---|---|---|---|---|---|---|---|---|---|---|---|
| | | P | R | F1 | P | R | F1 | P | R | F1 | |
| Traditional | SVM | 0.85 | 0.89 | 0.87 | 0.41 | 0.31 | 0.39 | 0.82 | 0.79 | 0.80 | 0.48 |
| BiLSTM | Word2vec | 0.89 | 0.93 | 0.91 | 0.65 | 0.53 | 0.59 | 0.88 | 0.81 | 0.83 | 0.66 |
| | fastText | 0.90 | 0.93 | 0.93 | 0.67 | 0.55 | 0.61 | 0.89 | 0.82 | 0.85 | 0.68 |
| | Self-learned | 0.89 | 0.92 | 0.90 | 0.61 | 0.49 | 0.55 | 0.84 | 0.77 | 0.79 | 0.62 |
| CNN | Word2vec | 0.91 | 0.93 | 0.92 | 0.66 | 0.55 | 0.61 | 0.89 | 0.83 | 0.85 | 0.69 |
| | fastText | 0.93 | 0.95 | 0.94 | 0.68 | 0.58 | 0.63 | 0.90 | 0.84 | 0.86 | 0.70 |
| | Self-learned | 0.89 | 0.93 | 0.91 | 0.62 | 0.50 | 0.56 | 0.85 | 0.78 | 0.80 | 0.64 |
| Transformers | mBERT | 0.94 | 0.90 | 0.92 | 0.68 | 0.58 | 0.64 | 0.90 | 0.85 | 0.87 | 0.71 |
| | XLM-R | 0.94 | 0.90 | 0.92 | 0.70 | 0.60 | 0.66 | 0.91 | 0.86 | 0.88 | 0.72 |
| | IndicBERT | 0.95 | 0.91 | 0.93 | 0.72 | 0.61 | 0.68 | 0.93 | 0.88 | 0.90 | **0.74** |
| Baseline | All TIN | 0.89 | 1.00 | 0.94 | 0.00 | 0.00 | 0.00 | 0.79 | 0.89 | 0.83 | 0.47 |
| | All UNT | 0.00 | 0.00 | 0.00 | 0.11 | 1.00 | 0.20 | 0.01 | 0.11 | 0.02 | 0.10 |

**Table 4** Results for offensive language categorization (level B). We report Precision (P), Recall (R), and F1 for each model/baseline on all classes (TIN, UNT), and weighted averages. Macro-F1 is also listed (best in bold).

| Type | Model | GRP | | | IND | | | OTH | | | Weighted | | | Macro F1 |
|---|---|---|---|---|---|---|---|---|---|---|---|---|---|---|
| | | P | R | F1 | P | R | F1 | P | R | F1 | P | R | F1 | |
| Traditional | SVM | 0.82 | 0.94 | 0.81 | 0.63 | 0.33 | 0.44 | 0.45 | 0.18 | 0.26 | 0.58 | 0.62 | 0.56 | 0.52 |
| BiLSTM | Word2vec | 0.82 | 0.94 | 0.81 | 0.67 | 0.38 | 0.48 | 0.49 | 0.25 | 0.31 | 0.62 | 0.65 | 0.60 | 0.56 |
| | fastText | 0.84 | 0.95 | 0.83 | 0.68 | 0.40 | 0.50 | 0.51 | 0.27 | 0.33 | 0.64 | 0.67 | 0.62 | 0.58 |
| | Self-learned | 0.84 | 0.95 | 0.84 | 0.69 | 0.41 | 0.51 | 0.51 | 0.28 | 0.34 | 0.65 | 0.67 | 0.62 | 0.58 |
| CNN | Word2vec | 0.84 | 0.96 | 0.83 | 0.69 | 0.40 | 0.50 | 0.51 | 0.27 | 0.33 | 0.64 | 0.67 | 0.62 | 0.58 |
| | fastText | 0.86 | 0.96 | 0.84 | 0.70 | 0.41 | 0.51 | 0.52 | 0.28 | 0.34 | 0.66 | 0.69 | 0.64 | 0.60 |
| | Self-learned | 0.84 | 0.96 | 0.83 | 0.69 | 0.40 | 0.50 | 0.51 | 0.27 | 0.33 | 0.64 | 0.67 | 0.62 | 0.58 |
| Transformers | mBERT | 0.86 | 0.97 | 0.85 | 0.72 | 0.43 | 0.53 | 0.54 | 0.32 | 0.38 | 0.68 | 0.70 | 0.65 | 0.62 |
| | XLM-R | 0.87 | 0.97 | 0.85 | 0.72 | 0.43 | 0.53 | 0.56 | 0.34 | 0.40 | 0.70 | 0.71 | 0.66 | 0.63 |
| | IndicBERT | 0.87 | 0.97 | 0.85 | 0.74 | 0.45 | 0.55 | 0.58 | 0.36 | 0.42 | 0.72 | 0.73 | 0.68 | **0.65** |
| Baseline | All GRP | 0.37 | 1.00 | 0.54 | 0.00 | 0.00 | 0.00 | 0.00 | 0.00 | 0.00 | 0.13 | 0.37 | 0.20 | 0.18 |
| | All IND | 0.00 | 0.00 | 0.00 | 0.47 | 1.00 | 0.64 | 0.00 | 0.00 | 0.00 | 0.22 | 0.47 | 0.30 | 0.21 |
| | All OTH | 0.00 | 0.00 | 0.00 | 0.00 | 0.00 | 0.00 | 0.16 | 1.00 | 0.28 | 0.03 | 0.16 | 0.05 | 0.09 |

**Table 5** Results for offense target identification (level C). We report Precision (P), Recall (R), and F1 for each model/baseline on all classes (GRP, IND, OTH), and weighted averages. Macro-F1 is also listed (best in bold).

### Offensive Language Target Identification

Here the models were trained to distinguish between three targets: a group (GRP), an individual (IND), or others (OTH). In Table 5, we can see that all the models achieved similar results, far surpassing the random baselines, with a slight performance edge for the transformer models. Similar to the previous levels, IndicBERT performed best in this level too providing 0.65 Macro F1 score.

# 5 Transfer-learning Experiments

The main idea of the methodology is that we train a classification model on a resource rich, typically English, using a cross-lingual language model, save the weights of the model and when we initialise the training process for Marathi, start with the saved weights from English. Previous work has shown that a similar transfer learning approach can improve the results for Arabic, Greek and Hindi [13, 16, 50]. We only experimented transfer-learning experiments with the transformer models as they provided better results than other embedding models in Section 4.

We first trained a transformer based classification model on a resource rich language. We used different resource rich languages for each level which we describe in the following sections. Then we save the weights of the transformer model as well as the softmax layer. We use this saved weights from the resource rich language to initialise the weights for Marathi.

### *Offensive Language Detection*

For the level A, we used several datasets as the resource rich language. As the first resource rich language we used English which can be considered as the language with highest resources for offensive language identification. We specifically used the OLID [14] level A tweets which is similar to the level A of MOLD 2.0. Also, in order to perform transfer learning from a closely-related language to Marathi, we utilised a Hindi dataset used in the HASOC 2020 shared task [51]. Both the English and Hindi datasets we used for transfer learning experiments contain Twitter data making them in-domain with respect to *MOLD 2.0*. The results are shown in Table 6.

| Language | Model | M F1 | W F1 |
|---|---|---|---|
| Hindi | XLM-R | 0.87 | 0.89 |
| English | XLM-R | 0.86 | 0.88 |
| - | IndicBERT | 0.85 | 0.88 |
| Hindi | IndicBERT | 0.85 | 0.88 |
| English | IndicBERT | 0.84 | 0.87 |
| - | XLM-R | 0.84 | 0.86 |
| Hindi | mBERT | 0.84 | 0.86 |
| English | mBERT | 0.83 | 0.85 |
| - | mBERT | 0.82 | 0.84 |

**Table 6** Transfer learning results for offensive language identification ordered by macro (M) F1 for MOLD 2.0. We also report weighted (W) F1 scores. For the comparison purpose, we also report the results for XLM-R, mBERT and IndicBERT when trained from scratch too.

As can be seen in the use of transfer learning substantially improved the monolingual results for mBERT and XLM-R. However, the IndicBERT model which performed best in the monolonigual experiments did not improve with the transfer learning approach. We believe that this can be due to the fact that the IndicBERT model is not cross-lingual. The best cross-lingual results were shown by the XLM-R model. From the two languages that we performed transfer learning, Hindi outperformed the results obtained using the English dataset suggesting that language similarity played a positive role in transfer learning.

*Categorization of Offensive Language*

For level B, we used the OLID level B as the initial task to train the transformer-based classification model. However, as far as we know, there are no datasets equivalent to MOLD 2.0 level B in related languages to Marathi such as Hindi and Bengali. Therefore, for level B, we only used English OLID level B as the initial task. The results are shown in Table 7.

| Language | Model     | M F1 | W F1 |
|----------|-----------|------|------|
| English  | XLM-R     | 0.75 | 0.91 |
| -        | IndicBERT | 0.74 | 0.90 |
| English  | mBERT     | 0.73 | 0.88 |
| English  | IndicBERT | 0.72 | 0.89 |
| -        | XLM-R     | 0.72 | 0.88 |
| -        | mBERT     | 0.71 | 0.87 |

**Table 7** Transfer learning results for categorisation of offensive language ordered by macro (M) F1 for MOLD 2.0. We also report weighted (W) F1 scores. For the comparison purpose, we also report the results for XLM-R, mBERT and IndicBERT when trained from scratch too.

As can be seen in the results, transfer learning improved the results for level B in XLM-R and mBERT. Similar to level A, IndicBERT performance was not improved with transfer learning. XLM-R with transfer learning provided the best results with 0.75 Macro F1 score.

*Offensive Language Target Identification*

As there are no equivalent datasets similar to level C in MOLD 2.0 in related languages, we only used OLID level C as the initial dataset.

| Language | Model     | M F1 | W F1 |
|----------|-----------|------|------|
| English  | XLM-R     | 0.74 | 0.90 |
| -        | IndicBERT | 0.74 | 0.89 |
| English  | IndicBERT | 0.73 | 0.88 |
| English  | mBERT     | 0.72 | 0.88 |
| -        | XLM-R     | 0.72 | 0.88 |
| -        | mBERT     | 0.71 | 0.87 |

**Table 8** Transfer learning results for offensive language target identification ordered by macro (M) F1 for MOLD 2.0. We also report weighted (W) F1 scores. For the comparison purpose, we also report the results for XLM-R, mBERT and IndicBERT when trained from scratch too.

As can be seen in Table 8 transfer learning improved the results of XLM-R and mBERT. However, in this level too, transfer learning did not improve the performance of IndicBERT. Overall, XLM-R with transfer learning provided the best result with 0.74 Macro F1 score.

# 6 SeMOLD - Semi-supervised Data Augmentation

For the semi-supervised experiments, we collected additional 8,000 Marathi, using the same methods described in Section 3. Rather than labelling them manually we followed a semi-supervised approach described to annotate SOLID [1]. We first selected the three best machine learning classifiers we had from Section 4: mBERT, XLM-R and IndicBERT. Then for each instance in the larger dataset, we saved the labels from each machine learning model. We release this larger dataset as SeMOLD: Semi-supervised Marathi Offensive Language Dataset. We use filtered SeMOLD instances to augment the training set. We only performed the data augmentation experiments for the transformer models.

### *Offensive Language Detection*

In the data augmentation process for level A, we augmented instances from SeMOLD, where at least two machine learning models predicted the same class in level A. For the level A, as can be seen in Table 9, when training with MOLD+SeMOLD, the results did not improve for the transformer models.

| Model     | MOLD | SeMOLD + MOLD |
|-----------|------|---------------|
| mBERT     | 0.82 | 0.82          |
| XLM-R     | 0.84 | 0.84          |
| IndicBERT | 0.85 | 0.84          |

**Table 9** Semi-supervised data augmentation results for offensive language identification in MOLD 2.0. We report the Macro F1 scores with the augmented data from SeMOLD in **SeMOLD + MOLD** column. For the comparison purpose, we report the results without data augmentation in **MOLD** column.

This is similar to the previous experiments in data augmentation [1] where the results do not improve when the machine learning classifier is already strong. We can assume that the transformer models are already well trained for MOLD and adding further instances to the training process would not improve the results for the transformer models.

### *Categorization of Offensive Language*

For the level B, the MOLD training set is smaller, and the task is also more complex than the level A. Therefore, the machine learning models can benefit from adding more data. As can be seen in Table 10, all of the transformer models improve with data augmentation from SeMOLD instances. IndicBERT model performed best with the data augmentation and provided 0.76 Macro F1 score.

| Model     | MOLD | SeMOLD + MOLD |
|-----------|------|---------------|
| mBERT     | 0.71 | 0.73          |
| XLM-R     | 0.72 | 0.74          |
| IndicBERT | 0.74 | 0.76          |

**Table 10** Semi-supervised data augmentation results for categorisation of offensive language in MOLD 2.0. We report the Macro F1 scores with the augmented data from SeMOLD in **SeMOLD + MOLD** column. For the comparison purpose, we report the results without data augmentation in **MOLD** column.

*Offensive Language Target Identification*

Finally for level C, the manually annotated OLID dataset is even smaller, and the number of classes increases from two to three. As can be seen in Table 11, all the models improve with the data augmentation process. IndicBERT model performed best after the data augmentation process scoring 0.68 Macro F1 score.

| Model     | MOLD | SeMOLD + MOLD |
|-----------|------|---------------|
| mBERT     | 0.62 | 0.64          |
| XLM-R     | 0.63 | 0.65          |
| IndicBERT | 0.65 | 0.68          |

**Table 11** Semi-supervised data augmentation results for offensive language target identification in MOLD 2.0. We report the Macro F1 scores with the augmented data from SeMOLD in **SeMOLD + MOLD** column. For the comparison purpose, we report the results without data augmentation in **MOLD** column.

# 7 Conclusion and Future Work

We presented a comprehensive evaluation of Marathi offensive language identification along with two new resources: MOLD 2.0 and SeMOLD. MOLD 2.0 contains over 3,600 tweets annotated with OLID's three-level annotation taxonomy making it the largest manually annotated Marathi offensive language dataset to date. SeMOLD is a larger dataset of 8,000 instances annotated with semi-supervised methods. Both these results open exciting new avenues for research on Marathi and other low-resource languages.

Our results show that it is possible to identify types and targets of offensive posts in Marathi with a relatively small size dataset (answering **RQ1**). With respect to **RQ2** we report that (2) the use of the larger dataset (SeMOLD) combined with MOLD 2.0 results in performance improvement particularly for levels B and C where less data is available in MOLD (answering **RQ2.1**); and (2) transfer learning techniques from both English and Hindi result in performance improvement for Marathi in the three tasks (identification, categorization, and target identification) (answering **RQ2.2**). We believe that these results shed light on offensive language identification applied to Marathi and low-resource languages as well, particular Indo-Aryan languages.

In future work, we would like to extend MOLD's annotation to a fine-grained token-level annotation. This would allow us to jointly model both instance label and token annotation as in MUDES [17]. Finally, we would like to use the knowledge and data obtained with our work on Marathi and expand it to closely-related Indo-Aryan languages such as Konkani.